\documentclass[letterpaper, 10 pt, conference]{ieeeconf}  
\usepackage{graphicx}
\usepackage{mdframed}
\usepackage{cite}
\usepackage{comment}
\usepackage{placeins}
\usepackage{url}
\usepackage{framed}
\usepackage{soul}
\usepackage{tcolorbox}
\usepackage{microtype} 
\usepackage{subcaption}

\newtcolorbox{mybox}{colback=orange!5!white,colframe=orange!75!black, top=0pt, bottom=0pt, left=3pt, right=3pt}
\makeatletter
\IEEEtriggercmd{\reset@font\normalfont\fontsize{6.9pt}{7.8pt}\selectfont}
\makeatother
\IEEEtriggeratref{1}

\IEEEoverridecommandlockouts                              

\title{\LARGE \bf
Understanding LLM Intervention Explanations in Multi-Party Human–Robot Interaction}

\author{Micol Spitale$^{1}$, Massimiliano Nigro$^{1}$, Emily Cross$^{2}$
\thanks{$^{1}$ Micol Spitale and Massimiliano Nigro are with the Department of Electronics, Information and Bioengineering, Politecnico di Milano, Milan, Italy. Corresponding author's email:
        {\tt\small micol.spitale@polimi.it}}%
\thanks{$^{2}$Emily Cross is with the GESS Department, ETH Zurich, Zurich, Switzerland}}
\begin{document}

\maketitle
\thispagestyle{empty}
\pagestyle{empty}

\begin{abstract}
Large Language Models (LLMs) are increasingly embedded in social robots to support natural group interactions, yet their role in complex multi-party settings remains underexplored. In particular, it is unclear how LLM-driven robots decide when and why to intervene in group conversations.
This paper investigates the intervention explanations generated by an LLM-based orchestrator in a multi-party interaction involving three human participants and two robots. We conducted a between-subjects study with 24 groups (66 university students), comparing a homogeneous condition (two robots with the same role, i.e., a mover) and a heterogeneous condition (two robots with different roles, i.e., a mover and an opposer).
At each conversational turn, the LLM orchestrator decided whether to intervene and generated a textual explanation of its decision. We performed a thematic analysis of 610 intervention explanations, identifying five recurring themes. Results show that explanations are facilitation-oriented, emphasizing agreement, participation, and interaction flow. While patterns remain stable across conditions, role differentiation emerges: the mover supports coordination, whereas the opposer drives goal-oriented  interventions.
These findings contribute to explainable AI by characterizing how LLM-driven systems justify intervention decisions in real-time, multi-party human–robot interaction.
\end{abstract}

\section{Introduction}
Research in Human--Robot Interaction (HRI) has increasingly explored the integration of Large Language Models (LLMs) into social robotic systems, enabling robots to participate in dynamic, naturalistic conversations with humans \cite{spitale2024appropriateness, zhang2023large, zhang2023largeiros, kim2024understanding, spitale2025past}. While prior work has shown promising results in dyadic HRI \cite{spitale2025vita, pinto2025designing}, multi-party settings with multiple humans and robots remain underexplored \cite{sebo2020robots, nigro2025social}. This gap is important because many real-world environments, such as classrooms, workplaces, and social settings, involve several human participants and may increasingly include multiple robotic agents. Such settings introduce additional interactional complexity, including coordination among robots \cite{yan2013survey}, distribution of conversational roles, turn-taking management, and collective decision-making \cite{parker2012decision}. Empirical studies of these complex multi-agent interactions remain limited, especially for LLM-powered autonomous robots.

At the same time, the growing use of LLMs in interactive systems raises questions about interpretability \cite{gantla2025exploring}. Although LLM-based agents can produce increasingly sophisticated conversational behavior \cite{abbo2025fast}, the way they decide whether to intervene, how to intervene, and what conversational goals to prioritize remains opaque \cite{zhao2024explainability, inoue2025llm, penzo2024llms}. In embodied settings, this opacity makes it difficult to assess whether system behavior is aligned or appropriate, particularly in socially sensitive domains \cite{raptis2025agentic, hadar2024assessing}. Despite broader interest in explainable AI, limited work has examined how LLM-driven systems justify intervention decisions in multi-party HRI.

To address this gap, we investigate intervention explanations generated by an LLM-based orchestrator in a multi-party interaction with three humans and two robots. At each turn, the orchestrator decided whether to intervene and which robot role to activate, generating a textual explanation for each decision. In this work, we use the term \textit{intervention explanations} to refer to these generated justifications. We avoid the term \textit{reasoning}, as we do not claim access to the model’s internal decision-making processes.

We analyzed data from a between-subjects study in which 24 groups, 66 university students in total, interacted with a Furhat\footnote{https://www.furhatrobotics.com/} and a Nao robot under two role configurations (study conditions). In the homogeneous (HOM) condition, both robots shared the same conversational role and acted as \textit{movers}, encouraging progress toward agreement. In the heterogeneous (HET) condition, one robot acted as a \textit{mover}, while the other acted as an \textit{opposer}, challenging proposals and introducing alternative perspectives. This design enabled us to examine whether intervention explanations remained stable across role configurations and whether differentiated robot roles exhibited distinct interactional functions.
Accordingly, this study addresses the following research questions:
\begin{itemize}
    \item \textbf{RQ1}: What types of intervention explanations characterize 
    LLM-driven decisions in multi-party HRI?
    \item \textbf{RQ2}: How does the social configuration of robots influence 
    the intervention explanations generated by the LLM orchestrator?
    Specifically, \textbf{RQ2a}) how do explanations differ between 
    HOM and HET configurations, and \textbf{RQ2b}) within 
    HET configurations, how do they differ between the 
    \textit{mover} and \textit{opposer} roles?
\end{itemize}
By addressing these questions, this work contributes to the study of LLM-powered social robots by characterizing intervention explanation patterns in a multi-human, multi-robot conversational setting. More broadly, it advances research on explainability in interactive AI by showing how generated justifications can be used to examine the behavior of embodied LLM-driven systems in real time.

\section{Related Work}

\subsection{Interpretability of LLMs in HRI}
The rapid integration of Large Language Models (LLMs) into interactive systems has intensified concerns about transparency and interpretability. Zhao et al. \cite{zhao2024explainability} survey of explainability techniques for LLMs, categorizing approaches into intrinsic, post-hoc, and interactive methods.
Within HRI, LLMs are increasingly used as high-level reasoning and dialogue engines \cite{abbo2025fast, addlesee2024multi}. Zhang et al. \cite{zhang2023large} review LLMs in HRI, highlighting their potential for flexible language understanding and generation, but also notingreliability, grounding, and transparency challenges. Similarly, Zhang and Soh \cite{zhang2023largeiros} propose using LLMs as zero-shot human models to simulate human reasoning in interactive scenarios, demonstrating their promise while raising questions about how such reasoning processes can be inspected and validated. Kim et al. \cite{kim2024understanding} further investigate user perceptions of LLM-powered robots, emphasizing the need to better understand how these systems generate responses and how humans interpret them. 

While these studies examine LLM-generated language quality, appropriateness, and user perception, less attention has been paid to analyzing the intervention explanation processes behind LLM-driven intervention decisions in multi-party settings. Our work contributes to this gap by empirically \textit{characterizing the explanations} accompanying real-time intervention decisions in multi-human, multi-robot interactions.

\subsection{Multi-party Human-Robot Interaction}
Research on robots in group settings predates the adoption of LLMs and has focused on coordination, role allocation, and social dynamics. Sebo et al. \cite{sebo2020robots} provide a comprehensive review of robots in groups and teams, identifying key challenges such as participation balance, role differentiation, and social influence. Nigro et al. \cite{nigro2025social} further outline computational challenges in social group HRI, including perception of group structure, turn-taking, addressee recognition, and coordination among multiple agents.
From a multi-robot systems perspective, decision-making and coordination have traditionally been framed as optimization or distributed control problems. Parker \cite{parker2012decision} and Yan et al. \cite{yan2013survey} survey methods for multi-robot coordination, focusing on task allocation, communication, and collective decision-making. However, these approaches typically emphasize efficiency and robustness rather than socially situated conversational reasoning.
With the introduction of LLMs, new challenges emerge in multi-party dialogue management. 
Addressee recognition and response selection are critical in group conversations. Inoue et al. \cite{inoue2025llm} propose a benchmark for evaluating LLMs on addressee recognition in multi-modal, multi-party dialogue, while Penzo et al. \cite{penzo2024llms} investigate whether LLMs struggle with “multi-party hangover” diagnosing limitations in tracking conversational context across multiple participants. 

These works highlight the cognitive and representational demands placed on LLMs in multi-party interaction. Our work advances multi-party HRI by analyzing intervention explanations in a multi-human, multi-robot setting, bridging \textit{LLM interpretability and multi-party HRI} research.

\section{Data Collection Study}

\begin{figure}
    \centering
    \includegraphics[width=\linewidth,trim={0 2cm 0 2.5cm}, clip]{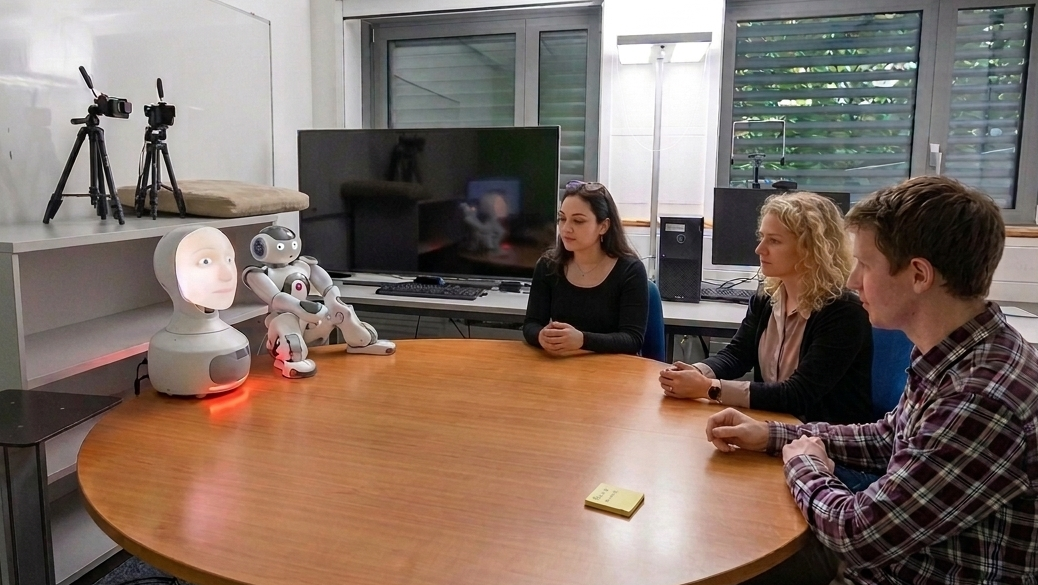}
    \caption{Experimental setup of the data collection study. Three human participants are seated around a round table, interacting with two robots (Nao robot and a Furhat robot). This image (originally with blurred faces) was edited with Gemini to generate fake faces of participant preserving their privacy.}
    \label{fig:setting}
\end{figure}

\begin{figure}
    \centering
    \includegraphics[width=\linewidth]{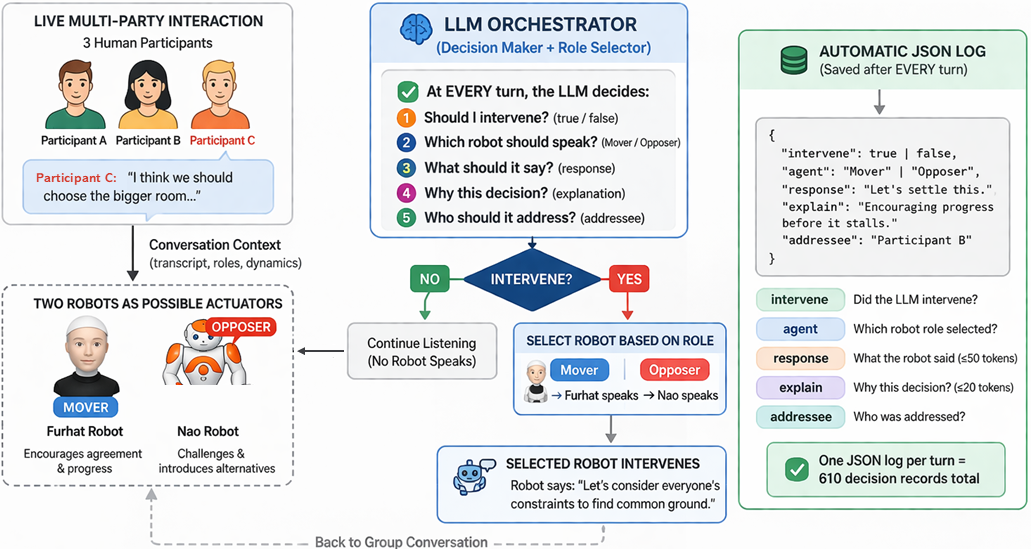}
    \caption{Overview of the LLM-orchestrated multi-party interaction. At each conversational turn, the LLM analyzes the group context and decides whether to intervene, which robot to activate (mover or opposer), what response to generate, and whom to address. Each decision is logged as a JSON.}
    \label{fig:interaction}
\end{figure}

Data were collected through a between-subjects study investigating multi-party interactions involving three human participants and two robots, namely a Furhat robot and a Nao robot (as depicted in Figure \ref{fig:setting}), that were controlled by a LLM orchestrator. Details on the prompts used to implement the orchestrator can be found in the paper's github folder\footnote{https://github.com/Massimilianonigro/MHRI-LLMIntervention}. Each session consisted of a group of two or three students engaging in a collaborative discussion task while interacting with two robotic agents.
The group’s task was to reach a consensus on how to allocate the rooms and their respective costs in a shared apartment with a total budget of 2,000 CHF. 
We decided not to set a time limit for the task, as the length of the conversation could vary depending on the LLM-generated responses. 
Instead, we fixed the number of turns in the group conversation to 30, which corresponded to approximately 15 minutes of interaction. 
Groups unable to reach a decision within 30 turns were considered not to have achieved consensus.
Two experimental conditions were implemented. In the homogeneous (HOM) condition, both robots were assigned the same conversational role: facilitating progress toward consensus (\textit{mover}). In the heterogeneous (HET) condition, the robots were assigned differentiated roles: one acted as a \textit{mover}, encouraging agreement and convergence, while the other acted as an \textit{opposer}, challenging proposals and introducing alternative perspectives. 
Robot roles were counterbalanced, and groups were randomly assigned to conditions.
The LLM orchestrator monitored the conversation in real time and made turn-by-turn decisions about whether to intervene, while also generating a textual explanation for each decision. 
At each decision point, the system logged a textual explanation alongside metadata on condition, role configuration, and whether an intervention occurred, as depicted in Figure \ref{fig:interaction}.
The final dataset comprised conversational logs from 24 groups of 66 university students, that resulted in 610 intervention decision instances of the LLM orchestrator. 
Each instance included (1) the conversational context, (2) the binary intervention decision (intervene vs. not intervene), and (3) the LLM-generated explanation underlying that decision. These logs formed the primary material for our thematic analysis.

\section{Data Analysis}
\subsection{Thematic Analysis}
For \textbf{RQ1}, we analyzed the distribution of explanations by conducting a thematic analysis and computing frequencies of both fine-grained codes and higher-level themes, in order to identify the most prevalent explanation patterns.

The coding scheme was developed iteratively, combining \textit{inductive and deductive strategies} \cite{clarke2017thematic}. An initial subset of the dataset was reviewed to identify recurring explanation patterns underlying intervention decisions. These patterns were translated into codes capturing distinct explanation types (e.g., encouraging agreement, balancing participation, proposing solutions).
Codes were iteratively refined, merged where appropriate, and operationally defined to ensure distinctiveness and mutual exclusivity.
The final coding scheme consisted of fine-grained codes, each accompanied by specific inclusion and exclusion criteria. 

One researcher  conducted the full coding of the dataset, and a second researcher independently reviewed 20\%, with an inter-rater agreement of 77\% and a Cohen's Kappa of 0.75. This process resulted in the identification of 22 codes, which were subsequently consolidated into five broader themes representing overarching intervention explanation orientations.
Discrepancies were resolved through consensus, after which the scheme was consistently applied across the dataset.
\subsection{Dialogue Analysis}
For \textbf{RQ2a}, we compared thematic distributions between HOM and HET conditions. We also compared intervention versus non-intervention frequencies across conditions. 
For \textbf{RQ2b}, we focused on the HET condition and analyzed how intervention explanations and generated utterances differed between the mover and opposer roles. We characterized LLM utterances using dialogue acts, i.e., functional conversation units \cite{bunt2005framework}, and addressees. Building on previous dialogue act taxonomies \cite{yu2021midas,core1997coding}, we developed a taxonomy tailored to social HRI, distinguishing between initiative acts (introducing new topics) and responsive acts (continuing existing ones). 

Initiative acts are comprised of questions and statements. Questions can be divided into factual questions, meant to seek specific information, to opinion questions, that elicit subjective responses, and small talk, meant to follow social conventions. Statements allow for initiating new conversation topics or adhering to social conventions through declarative utterances.
Responsive acts continue existing conversation threads and are subdivided into three categories: acknowledgments, follow-ups, and answers. Acknowledgements address a previous point in the conversation through comments that affirm without elaborating or opinions that add new information to the exchange. Follow-ups build on previous conversations by posing questions that extend the dialogue. Answers respond directly to prior questions. We divide answers into positive, negative, and generic other answers.
One researcher manually annotated robot utterances for dialogue acts, and coded addressees as either the whole group or a single individual.  
\section{Findings}

\begin{figure*}[t]

    \begin{subfigure}{0.48\textwidth}
        \centering
        \includegraphics[width=\linewidth]{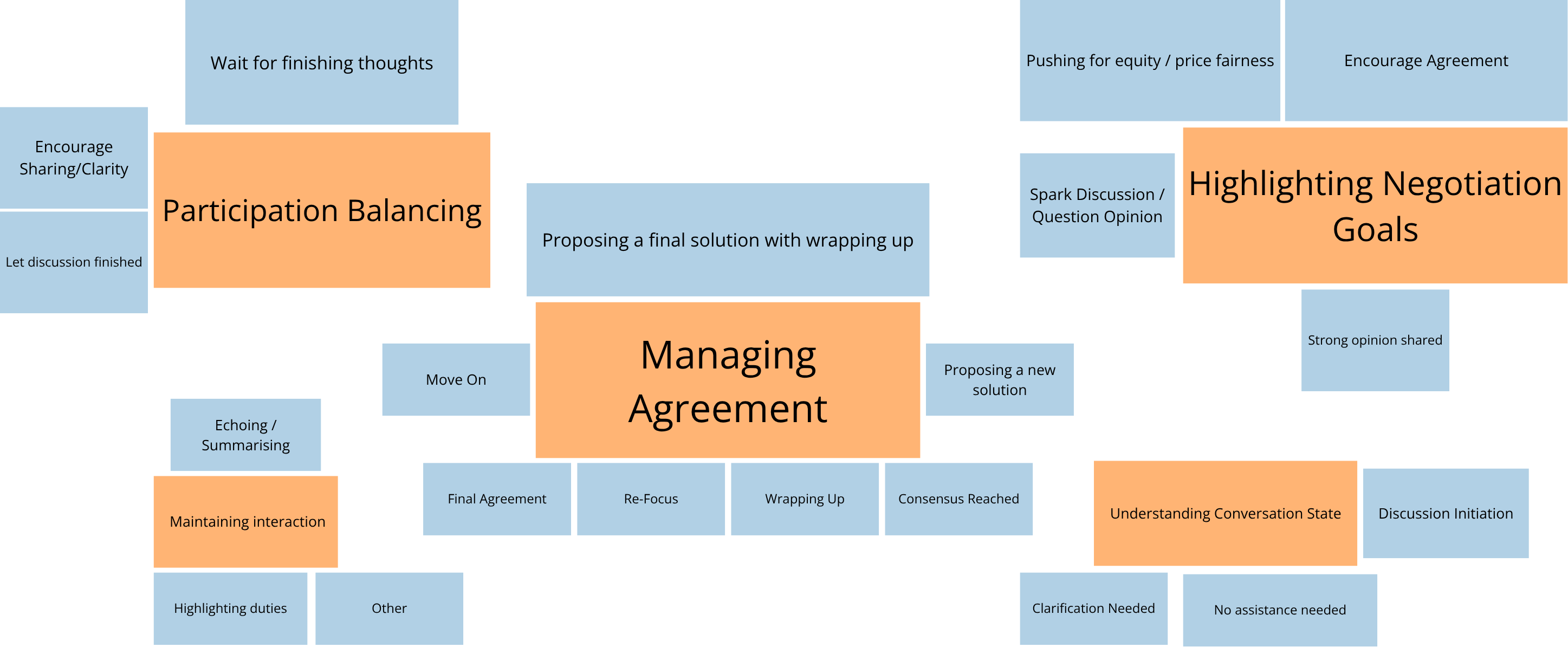}
        \caption{Theme distribution in the HET condition}
        \label{fig:theme_distr_het}
    \end{subfigure}
    \hfill
    \begin{subfigure}{0.48\textwidth}
        \centering
        \includegraphics[width=\linewidth]{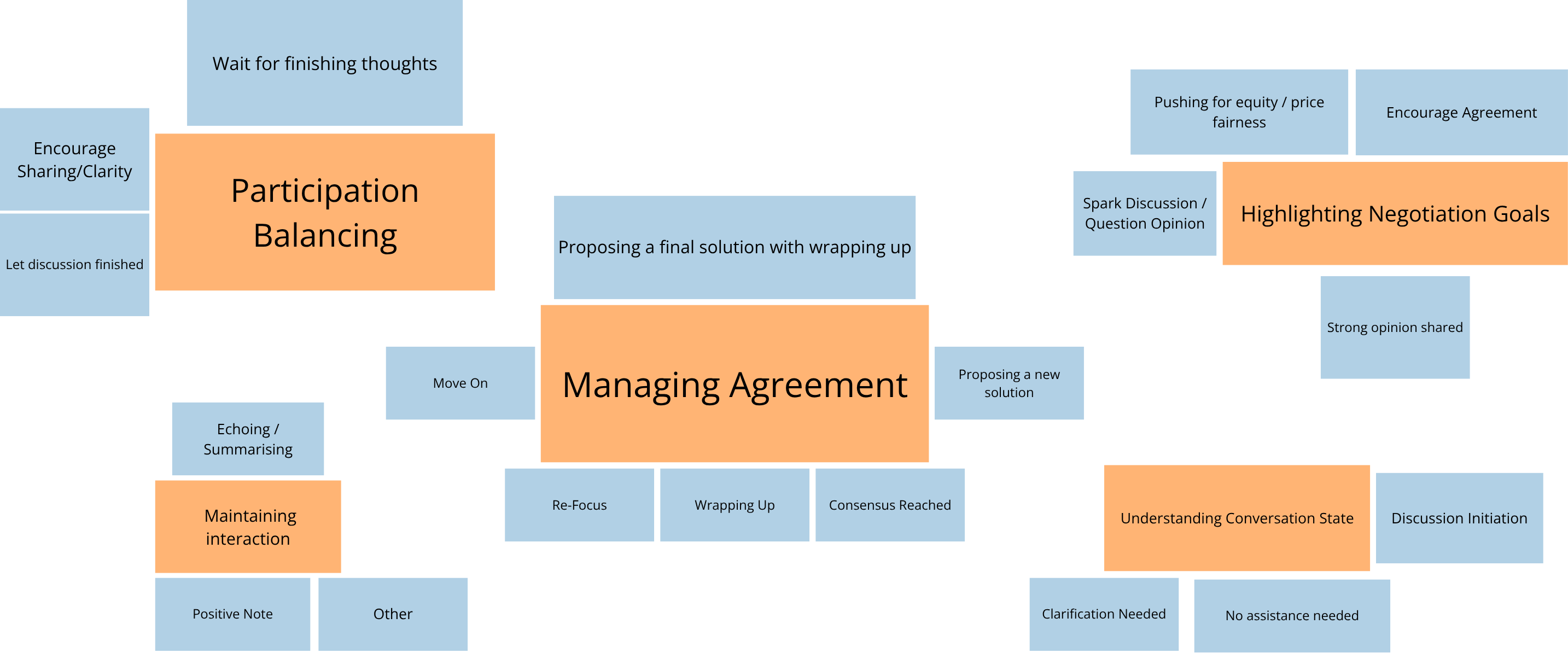}
        \caption{Theme distribution in the HOM condition}
        \label{fig:theme_distr_hom}
    \end{subfigure}
\end{figure*}
\subsection{What types of intervention explanations characterize LLM-driven decisions in multi-party HRI? (RQ1)}

The thematic analysis included 610 coded instances distributed across 22 fine-grained codes, which were grouped into five broader themes, namely ``Managing Agreement" ($n\!=\!202$), ``Participation Balancing" ($n\!=\!167$), ``Highlighting Negotiation Goals"  ($n\!=\!152$), ``Maintaining Interaction" ($n\!=\!59$), and  ``Understanding Conversation State" ($n\!=\!30$). At the code level, the most frequent category was ``proposing a final solution with wrapping up'' ($n\!=\!106$), followed by ``wait for finishing thoughts'' ($n\!=\!91$) and ``encourage sharing / clarity''($n\!=\!64$). Other relatively frequent codes included ``spark discussion / question opinion'' ($n\!=\!55$), ``pushing for equity / price fairness'' ($n\!=\!44$) and ``encourage agreement'' ($n\!=\!40$). Less frequent codes included and ``echoing/summarizing'' ($n\!=\!26$), ``proposing a new solution'' (i.e., not final, but still in the middle of the discussion) ($n\!=\!25$), ``move on'' ($n\!=\!23$), ``discussion initiation'' ($n\!=\!22$) and ``other'' ($n\!=\!22$). All remaining codes appeared only occasionally. These fine-grained codes were consolidated into five broader themes, described below.

\paragraph{Managing Agreement} ``Managing Agreement" was the most frequent theme ($n\!=\!202$). It was mainly composed of ``proposing a final solution with wrapping up'', followed by ``proposing a new solution'', ``move on'', ``wrapping up'',``re-focus'', ``consensus reached'' and ``final agreement''. 
This theme reflected efforts to move the discussion toward closure by consolidating proposals, marking convergence, and signaling readiness to conclude, indicating an orientation toward \textbf{progress and resolution}. 
\paragraph{Participation Balancing} ``Participation Balancing" was the second most frequent theme for 167 instances and was largely driven by ``wait for finishing thoughts''. Additional contributions included ``encourage sharing/clarity'' and ``let discussion finished''.
This theme reflects a concern with regulating conversational pacing and participation equity. The explanations justified restraint, allowing speakers to complete their contributions and reducing interruptions. This suggests that \textbf{sensitivity to turn-taking norms and procedural fairness} was a central aspect of the interaction.
\paragraph{Highlighting Negotiation Goals} ``Highlighting Negotiation Goals" accounted for 152 instances. It was primarily composed of ``spark discussion / question opinion'', ``pushing for equity / price fairness'', and ``encourage agreement'' with smaller contributions from ``proposing schedule / rotation'' and ``strong opinion shared''.
This theme captures goal-directed explanations aimed at influencing the group's decision process. Explanations revolved around fairness and distributive equity, indicating that \textbf{equity considerations were central to how the orchestrator framed negotiation goals}. Another portion focused on eliciting participant opinions or steering the discussion toward underexplored topics in the negotiation.
\paragraph{Maintaining Interaction} ``Maintaining Interaction" totaled for 59 instances and was driven primarily by the code ``echoing/summarising'' and ``other'' with smaller contributions from ``positive note''. 
These explanations reflected structural, non-strategic contributions that sustained the conversational environment without persuading participants or regulating participation, suggesting the orchestrator was often oriented toward \textbf{maintaining the interaction space} rather than directly shaping the substance of the discussion.
A few isolated instances within this theme referred to duties or responsibilities grouped by the code ``highlighting duties''. 
\paragraph{Understanding Conversation State} ``Understanding Conversation State" was the least frequent theme ($n\!=\!30$). It mainly included ``discussion initiation'' , together with ``no assistance needed'' and ``clarification needed''. These intervention explanations focused on the \textbf{state of the conversational interaction} itself rather than the content of the task. They reflected moments in which the orchestrator attended to how the conversation was unfolding by initiating discussion, clarifying the state of the exchange, or signaling that no intervention was needed. Although less frequent, these instances show that the orchestrator occasionally adopted a more meta-level orientation toward the conversation.

\begin{mybox}
The findings indicate that intervention explanations alternated between promoting group consensus, ensuring equal participation, and encouraging discussion of relevant topics. These results suggest the system effectively addressed key challenges in the negotiation process of multi-party HRI.
\end{mybox}

\subsection{How do intervention explanations differ between homogeneous and heterogeneous robot role configurations? (RQ2a)}

In the HET condition, the distribution of themes was: ``Managing Agreement'' ($n\!=\!107$), ``Highlighting Negotiation Goals'' ($n\!=\!92$), ``Participation Balancing'' ($n\!=\!83$), ``Maintaining Interaction'' ($n\!=\!21$) and ``Understanding Conversation State'' ($n \!=\!16$). 
In the HOM condition, the distribution was: ``Managing Agreement'' ($n\!=\!95$), ``Participation Balancing'' ($n\!=\!84$), ``Highlighting Negotiation Goals'' ($n\!=\!60$), ``Maintaining Interaction'' ($n\!=\!38$), ``Understanding Conversation State'' ($n\!=\!14$).
``Managing Agreement'' was the most frequent theme, indicating that most intervention explanations pushed towards an agreement in the negotiation. This pattern was consistent between HET ($n\!=\!107$) and HOM ($n\!=\!95$).
Similarly, ``Participation Balancing'' appeared at similar levels in HET ($n\!=\!83$) and HOM ($n\!=\!84$), suggesting that ensuring equal participation in the discussion was a stable function of the orchestrator regardless of role configuration.
Slight differences were observed for ``Highlighting Negotiation Goals'', which appeared more frequently in HET ($n\!=\!92$) compared to the HOM ($n\!=\!60$). Although these differences suggest a tendency toward more goal-oriented intervention explanations when roles are differentiated, these variations were not statistically significant. 
Finally, ``Understanding Conversation State'' occurred at similar frequencies HOM condition ($n\!=\!18$) than in HET ($n\!=\!13$).

We also examined if the likelihood of intervention differed between conditions. In the HOM condition, the orchestrator intervened 226 times and refrained from intervening 65 times. In the HET condition, it intervened 241 times and did not intervene 78 times. 
This indicates that role configuration did not affect the orchestrator's propensity to intervene.

\begin{mybox}
Altering robot role configuration (HET vs. HOM) did not meaningfully change the LLM orchestrator’s intervention explanation patterns or its likelihood to intervene. 
\end{mybox}

\begin{figure*}[htb!]
\centering
\begin{minipage}{0.45\textwidth}
    \includegraphics[width=\linewidth, trim={0 0 0 7cm},clip]{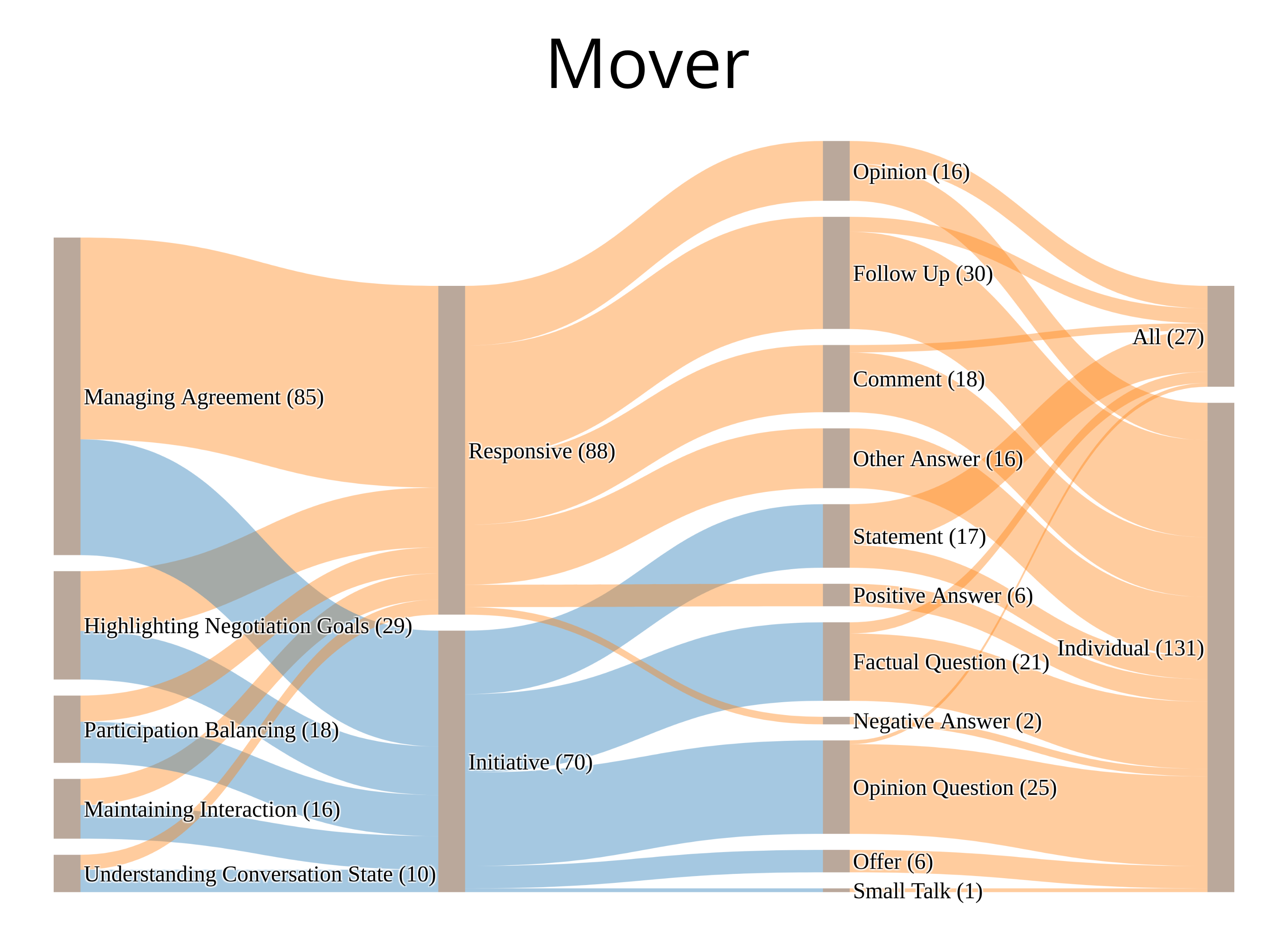}
    \caption{Sankey Diagram depicting dialogue acts and addressees of the \textit{mover}'s utterances}
    \label{fig:flow_mover}
\end{minipage}
\hfill
\begin{minipage}{0.45\textwidth}
    \includegraphics[width=\linewidth,trim={0 0 0 7cm},clip]{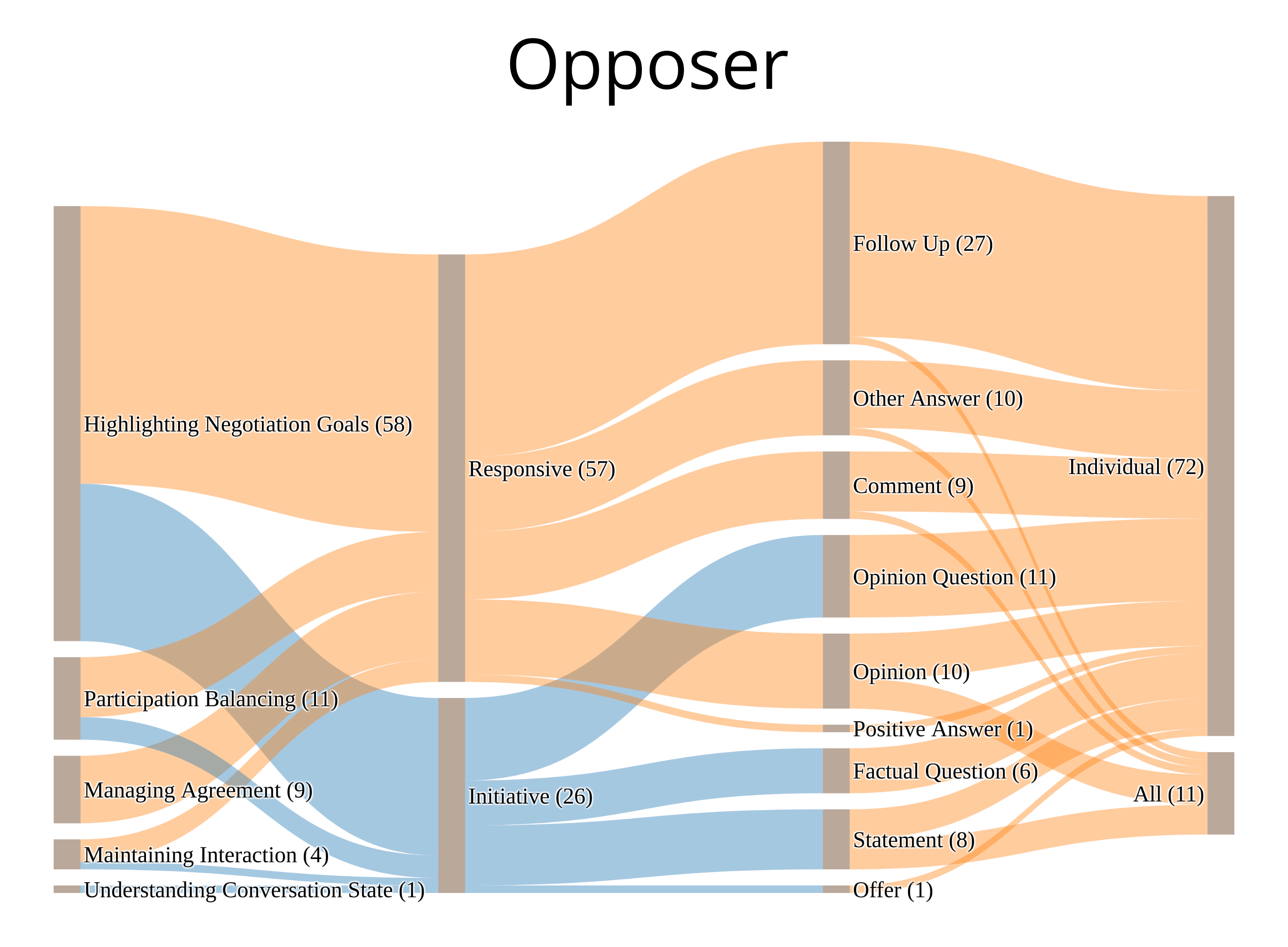}
    \caption{Sankey Diagram depicting dialogue acts and addressees of the \textit{opposer}'s utterances}
    \label{fig:flow_opposer}
\end{minipage}    
\end{figure*}

\subsection{In heterogeneous configurations, how do intervention explanations differ between distinct robot roles (mover vs. opposer)? (RQ2b)}

We first analyzed the LLM orchestrator’s intervention explanations and its robot selection at each conversational turn.
The \textit{mover} was selected more often than the  \textit{opposer} (158 vs. 83), particularly for ``Managing Agreement'' (85 vs. 9), ``Participation Balancing'' (18 vs. 11), and ``Maintaining Interaction'' (16 vs. 4).  The \textit{opposer} was preferred only for ``Highlighting Negotiation Goals'' (29 vs. 58), with all differences statistically significant according to two-sided Mann--Whitney U tests ($p < .001$).

We further examined how intervention explanation themes are reflected in the \textit{dialogue acts} produced by each robot role (see Figures \ref{fig:flow_mover}, and \ref{fig:flow_opposer}). 
Both roles predominantly generated responsive utterances. However, this tendency was significantly stronger for the \textit{opposer} (57/83 utterances) than for the \textit{mover} (88/158 utterances), as confirmed by a Mann--Whitney U test ($p \!=\! .0494$).
When examined by theme, the \textit{mover} displayed a relatively balanced distribution across dialogue act categories, combining both initiative and responsive contributions. In contrast, the \textit{opposer} showed a clear skew toward responsive acts, particularly in explanations related to ``Managing Agreement" and ``Highlighting Negotiation Goal".
This pattern is further reflected in the use of the ``Follow-Up'' dialogue act (i.e., questions that extend an existing topic). The \textit{opposer} used follow-up questions in 32.5\% of its utterances (27/83), compared to 20\% for the \textit{mover} (30/153).

\begin{mybox}
The LLM orchestrator assigned complementary roles to the robots: the \textit{opposer} primarily reacted by challenging participants’ views, while the \textit{mover} both responded and introduced new topics to guide the conversation. The \textit{mover} was preferred for facilitation (e.g., balancing participation, fostering agreement), whereas the \textit{opposer} was selected to stimulate focused, goal-oriented discussion.
\end{mybox}

\section{Discussion}
This study analysed intervention explanations of an LLM-orchestrator in a multi-party multi-robot HRI setting. 

A central finding is that the LLM orchestrator largely operates as a facilitator of the group process rather than as an active contributor to decision content (\textbf{RQ1}). Its behavior is strongly oriented toward managing agreement, balancing participation, and maintaining progress, which aligns with prior work emphasizing structured facilitation and equitable participation in multi-party interaction \cite{gillet2022learning}. This raises the question of whether such facilitation-heavy behavior may limit its capacity to meaningfully shape outcomes.
Agreement-oriented interventions suggest a bias toward consensus-seeking indicating that the orchestrator tends to steer discussions toward convergence. While this supports task completion it potentially comes at the expense of deeper deliberation. 
In this context, the recurring emphasis on fairness—particularly in goal-directed explanations—appears to function as an implicit normative anchor. Framing interventions around fairness and equity may enhance interpretability by leveraging shared social norms, but it also raises concerns about how such values are internalized by LLMs and how they might subtly influence group dynamics \cite{wang2024ali}.
Notably, meta-level awareness of the conversation (e.g., understanding the interaction state) and procedural support (e.g., maintaining interaction continuity) are present but less dominant. 
Yet, their presence is important: even infrequent meta-level and procedural explanations contribute to making the system’s behavior more legible and predictable \cite{dass2025improving, sharma2025adaptive}.

Contrary to expectations, explanation distributions remained stable across conditions (\textbf{RQ2a}), with no statistically significant differences observed. This likely reflects the orchestrator's preference for the mover over the opposer even in HET (153 vs. 83 instances), resulting in the two conditions having similar interaction patterns. From a design perspective, this stability may be advantageous as consistent patterns support predictability in multi-agent systems \cite{zhu2025dexter}, though it suggests that assigning different conversational roles may not be sufficient to induce variation in explanation strategies, and more explicit mechanisms may be required.

Even though, in the HET condition (\textbf{RQ2b}) the orchestrator still assigns different functions to the two robots. The \textit{mover} is primarily associated with coordination-oriented behaviors—such as managing agreement, balancing participation, and maintaining interaction—whereas the \textit{opposer} is more often deployed in goal-oriented interventions that challenge and refine the discussion. Our findings suggest that the LLM orchestrator strategically leverages role-specific interaction patterns.
This differentiation becomes more evident at the level of enacted behavior than in the high-level distribution of explanation themes. While both roles remain largely reactive, the \textit{opposer} more consistently builds on and challenges participants’ contributions, whereas the \textit{mover} adopts a hybrid strategy, combining responsiveness with initiative to guide the discussion. This indicates that role specialization is expressed less through what is explained and more through how interventions are performed in interaction.
These findings align with prior work in multi-party HRI showing that robots can shape group dynamics through differentiated interaction strategies \cite{sebo2020robots, traeger2020robots}, though whether role-based orchestration enables  diverse behaviors or merely reinforces a facilitation-opposition divide remains open.

Overall, our findings show that LLM-generated intervention explanations in multi-party HRI are predominantly fairness-oriented, and consistent across configurations. Yet differences emerge in how robot assigned to different roles behave, highlighting the importance of jointly analyzing explanations and conversational actions when evaluating LLM-driven systems in socially complex environments.

\section{Limitations and Future Works}
Several limitations should be acknowledged. First, although we characterized robot behavior through LLM-generated explanations, we did not evaluate whether explanations are understandable to users or improve perceived system transparency;  future work should address this gap.
Second, while we categorized the reasoning behind robot interventions, we did not link these behaviors to objective measures of the conversation (e.g., task success rate). As a result, questions such as how robot interventions influence group behavior, or whether different interaction patterns emerge in groups with different outcomes, remain open.
Finally, this study remains exploratory and does not establish strong theoretical claims; future research should test generalizability across tasks, environments, and system configurations, and develop models of orchestrator decision-making in multi-party HRI.

\section{Conclusion}
This work investigated the interpretability of LLM-driven intervention decisions in a multi-human, multi-robot interaction setting. Through thematic analysis of 610 decision explanations, we identified five overarching intervention explanation orientations, dominated by  progress monitoring, participation balancing and fairness-driven steering. These intervention explanation patterns remained stable across HOM and HET robot role configurations, with differences in robot role being reflected more in the enacted behaviour of the robots.
By systematically characterizing real-time intervention explanations, this study contributes to bridging explainable AI and multi-party HRI. It demonstrates that interpretability in embodied LLM systems can be empirically examined through the analysis of decision justifications, offering a concrete pathway toward more transparent and accountable multi-robot conversational orchestration.

\section*{Acknowledgment}
\noindent
\footnotesize
\textbf{Fundings:} M. Spitale has been funded by IDEA League Fellowship. 
\textbf{Use of GenAI:} ChatGPT was used during the writing phase to help with rephrasing and grammar correction. 




\bibliographystyle{IEEEtran}
\bibliography{bibliography}

\end{document}